\documentclass[a4paper]{article}

\usepackage{INTERSPEECH2022}
\usepackage{url}
\usepackage{xcolor}

\title{Exploring Capabilities of Monolingual Audio Transformers using Large Datasets in Automatic Speech Recognition of Czech}
\name{Jan Lehečka, Jan Švec, Aleš Pražák, Josef V. Psutka}
\address{Department of Cybernetics, University of West Bohemia Pilsen, Czech Republic}
\email{\{jlehecka,honzas,aprazak,psutka\_j\}@kky.zcu.cz}

\begin{document}

\maketitle

\begin{abstract}
In this paper, we present our progress in pretraining Czech monolingual audio transformers from a large dataset containing more than 80 thousand hours of unlabeled speech, and subsequently fine-tuning the model on automatic speech recognition tasks using a combination of in-domain data and almost 6 thousand hours of out-of-domain transcribed speech. 
We are presenting a large palette of experiments with various fine-tuning setups evaluated on two public datasets (CommonVoice and VoxPopuli) and one extremely challenging dataset from the MALACH project.
Our results show that monolingual Wav2Vec 2.0 models are robust ASR systems, which can take advantage of large labeled and unlabeled datasets and successfully compete with state-of-the-art LVCSR systems. Moreover, 
Wav2Vec models proved to be good zero-shot learners when no training data are available for the target ASR task.

%our general transformer-based ASR system can be used as a base model to recognize speech in the arbitrary domain and -- if small in-domain training data are available -- easily fine-tuned to the target domain while transferring the knowledge from the general ASR system.

\end{abstract}

\section{Introduction}
Self-supervised models recently became a very popular alternative to large vocabulary continuous speech recognition (LVCSR) systems in automatic speech recognition (ASR) tasks. They can learn contextualized speech representations from large-scale unlabeled audio datasets (pretraining self-supervised phase), and consequently employ the knowledge in the ASR training from labeled data (fine-tuning supervised phase). One of the most studied self-supervised ASR model architectures is Wav2Vec 2.0 \cite{baevski2020wav2vec}. It is a deep neural network pretrained to reconstruct the corrupted signals. The input raw audio signal is processed by a multi-layer convolutional neural network into a sequence of latent-speech representations which are fed into a multi-layer Transformer \cite{vaswani2017attention}. The output of the Transformer is a sequence of frame-level contextualized speech representations which are then processed by the connectionist temporal classification (CTC) layer \cite{graves2006connectionist,Baevski2020EffectivenessOS} decoding the most probable sequence of graphemes.

In contrast with standard LVCSR systems, these end-to-end approaches alleviate the need for word pronunciation modeling and do not require any alignment of data.

The main focus of this paper is to explore the capabilities of monolingual Wav2Vec-based models, their ability for zero-shot transfer learning, performance dependency on the training metaparameters, scaling for large ASR datasets and additional language models (LMs). We also compare the end-to-end models with state-of-the-art hybrid DNN-HMM LVCSR systems.

\section{Related work}
Monolingual Wav2Vec models for languages other than English are very rare \cite{evain2021task}. For languages like Czech, there are none. 
However, there are many multilingual pretrained models of sizes from large \cite{conneau2020unsupervised} to extremely large \cite{babu2021xls}. These models include also Czech. 
The common practice is to adopt a multilingual pretrained model and fine-tune it on the labeled ASR data from the target language. 
Since we had access to large unlabeled datasets and were not satisfied with results from multilingual models, we decided to pretrain our own monolingual Wav2Vec model from scratch and released it to the public. We are not aware of any similar model for Czech mentioned in the literature.

\section{Pretraining}
Self-supervised audio transformers are known to scale well with the size of pretraining data, even with extremely huge datasets \cite{babu2021xls}. Hence, we tried to gather as much public and in-house unlabeled audio data as possible. Together, we were able to collect more than 80 thousand hours of Czech speech. We are not aware of any similar collection of Czech speech data at this scale mentioned in the literature so far. The collection includes recordings from radio (22k hours), unlabeled data from VoxPopuli dataset \cite{wang-etal-2021-voxpopuli} (18.7k hours), TV shows (15k hours), shadow speakers (12k hours), sports (5k hours), telephone data (2k hours), and a smaller amount of data from several other domains. We used also raw audio files from all speech recognition datasets (see Sec. \ref{sec:ft}). 

Since the feature extraction of the input signal is limited by the memory of GPUs in use, we sliced all records not to exceed 30\,s, which we found to be a reasonable input size for batching. 

We followed the same pretraining steps as for the base Wav2Vec 2.0 model in \cite{baevski2020wav2vec}. We pretrained the model for 400 thousand steps with a batch size of about 1.6 hours, corresponding to more than 11 epochs over the dataset. 
%The pretraining took about two weeks on a machine with 4x NVIDIA A100 GPU. 
We released our pretrained model under the nickname \emph{ClTRUS} (abbreviation for \textbf{C}zech \textbf{l}anguage \textbf{TR}ransformer from \textbf{U}nlabeled \textbf{S}peech) for public non-commercial use\footnote{Available at \url{https://huggingface.co/fav-kky/wav2vec2-base-cs-80k-ClTRUS}}.

\section{Fine-tuning}
\label{sec:ft}
We prepared all training and development speech recognition data consistently for all datasets. We sliced long training audio signals on speech pauses not to exceed the length of 30\,s,
longer utterances were discarded.
%Due to the memory limit of GPUs, we discarded all utterances longer than 30\,s. 
We removed non-speech events and punctuation from the transcripts and mapped texts into lowercase. For each dataset, we carefully analyzed transcripts and fixed any data-specific deviation.

If not stated otherwise, we fine-tuned all models with the same setting as the base model in \cite{baevski2020wav2vec} using \texttt{Fairseq} tool\footnote{\url{https://github.com/pytorch/fairseq}}.

%\section{Processing long audio files}
%Okénkování - window, overlap, možná obrázek -> nechat na TSD

\subsection{Speech recognition datasets}
\label{sec:data}
We were experimenting with three Czech speech recognition datasets. Basic statistics are shown in Tab.~\ref{tab:ASRstats}. 

\setlength{\tabcolsep}{0.25em}
{\renewcommand{\arraystretch}{1.1}%
\begin{table}[htb]
  \caption{Speech recognition datasets. We are showing the number of hours, the number of words in transcripts (in thousands), and the average length of records (in seconds).}
  \label{tab:ASRstats}
  \centering
  \begin{tabular}{lrrrcrrrcrrr}
    \toprule
     & \multicolumn{3}{c}{CommonVoice} & & \multicolumn{3}{c}{VoxPopuli} & & \multicolumn{3}{c}{MALACH}\\
    \cline{2-4} \cline{6-8} \cline{10-12} 
     & train & dev & test & & train & dev & test & & train & dev & test \\
    \midrule
    
    \textit{\# hours} & 32.2 & 8.1 & 8.1 & & 52.3 &  3.0 & 3.1 & & 87.2 & 19.2 &  9.0 \\
    \textit{\# words} & 183 & 46 & 45 & & 404 &  23 & 23 & & 615 & 137 &  63 \\
    \textit{avg-len}  &  4.1 & 4.6 & 4.5 & & 10.2 & 10.0 & 9.9 & & 24.1 & 24.1 & 10.6 \\
  
    \bottomrule
  \end{tabular}
\end{table}
}

The \textbf{CommonVoice} dataset is a Czech portion of crowdsourced project Mozilla Common Voice \cite{commonvoice:2020}. We used corpus version 7.0 containing 49 hours of validated speech. We decided to keep also sentences reported as \textit{offensive} and \textit{difficult pronunciation} in our training data. All other reported sentences (e.g. \textit{incomplete transcription}, \textit{different language} etc.) were ignored.

The \textbf{VoxPopuli} dataset \cite{wang-etal-2021-voxpopuli} is a large-scale multilingual speech corpus collected from 2009-2020 European Parliament event recordings. The Czech portion contains $18.7$ thousand unlabeled hours and 62 hours with transcription. We ignored all train/dev records without the raw transcription, decreasing the amount of transcribed data to 58.4 hours.

The \textbf{MALACH} data\footnote{\url{https://malach.umiacs.umd.edu}} is a subset of the USC Shoah Foundation Visual History Archive of digitized interviews in 32 languages from 52,000 survivors, liberators, rescuers, and witnesses of the Nazi Holocaust. We used the Czech portion \cite{MALACHcz} which is known to be a very difficult and challenging speech recognition task due to the strong emotional and heavily accented speech of Holocaust survivors \cite{MALACH1,MALACH2}. 

To scale the fine-tuning up, we used also a large additional in-house dataset, denoted as \textbf{Extra}, containing almost 6 thousand hours of transcribed Czech speech from various domains. The dataset includes radio shows records (3.9 thousand hours), TV sports recordings (645 hours), telephone recordings (440 hours), and several other domains. The transcripts contain 43 million words.

\section{Decoding}
We studied two different decoding setups: (1) connectionist temporal classification (CTC) \cite{graves2006connectionist}, which is the training loss we used during fine-tuning of the models, and (2) CTC beam search decoder with an LM. CTC is an alignment-free method for grouping audio frames belonging to the same output symbol in order to convert a sequence of audio frames into a much shorter sequence of characters. Thus, Wav2Vec with the CTC is a grapheme-based lexicon-free speech recognizer without any language constraints. The only orthography-related knowledge the model could learn is the training transcripts fed in during the fine-tuning. Incorporating an LM into the CTC beam search decoder usually improves the speech recognition accuracy by bringing useful language information into the decoding process and penalizing improbable outputs. On the other hand, it is a step back from the idea of an end-to-end recognizer introducing once again problems known from LVCSR systems, such as out-of-vocabulary words and the low-relevance or low-quality LMs. 

For our experiments, we prepared 3 different word-based n-gram LMs. (1) \textbf{LM-ASRSpec} is a model trained specifically for each speech recognition dataset only from in-domain training transcripts. (2) \textbf{LM-ASRAll} was trained from transcripts from all speech recognition training data described in Sec. \ref{sec:data}, including the Extra dataset. It is a general domain-independent LM trained from ASR transcripts with  44 million words. (3) \textbf{LM-C5} is an LM trained from Czech Colossal Clean Crawled Corpus (C5) \cite{FERNET} which is a huge collection of cleaned and deduplicated web pages from Common Crawl project\footnote{\url{https://commoncrawl.org}}. Since this text corpus contains almost 13 billion words (93 GB of cleaned text), we pruned all unigrams with counts lower than 10 and higher-order n-grams with counts lower than 100.

We used implementation from \texttt{Transformers} \cite{wolf-etal-2020-transformers} for CTC decoding and \texttt{pyctcdecode}\footnote{\url{https://github.com/kensho-technologies/pyctcdecode}} decoder for CTC beam search decoder with n-gram LM. To train LMs, we used \texttt{KenLM} \cite{heafield2011kenlm}  and to work with models of practical sizes, we limited the maximum order of models to 4-grams. We trained all LMs in lowercase. The sizes of LM vocabularies were between 17 and 22 thousand (LM-ASRSpec), 329 thousand (LM-ASRAll), and 4.8 million words (LM-C5). 
%The great advantage of Wav2Vec models over LVCSR systems is that there is no need to specify a pronunciation lexicon for words in LM.
%We were experimenting also with other n-gram orders but we observed only expected behavior. 

\section{Experiments}
We evaluated speech recognition systems on test splits of individual datasets and compared word error rates (WER).
%or character error rate (CER) in the case of grapheme-based models.
During the evaluation, we mapped reference texts into lowercase and ignored all punctuation.

\subsection{CTC beam search decoder with LM}
To get a picture of how the recognition is improved when adding LM into the decoder, we prepared data-specific Wav2Vec models (denoted as W2V-ASRSpec) by fine-tuning the pretrained model only on in-domain (single-dataset) training data.
Then, we evaluated the models with different LMs in the decoder. From results tabulated in Tab.~\ref{tab:CTCLM}, we can see that adding small data-specific LM improves the recognition, adding domain-independent LM from all ASR transcripts further improves the recognition, and switching to large-scale LM trained from Common Crawl could be also beneficial, especially for domains with specific words mentioned several times somewhere on the Internet (e.g. names of politicians for VoxPopuli or geographical names for MALACH).

\setlength{\tabcolsep}{0.3em}
{\renewcommand{\arraystretch}{1.0}%
\begin{table}[htb]
  \caption{WER $[\%]$ for CTC beam search decoder with different LMs evaluated on three Czech datasets. }
  \label{tab:CTCLM}
  \centering
  \begin{tabular}{lrrr}
    \toprule
     & \hspace{-3em} CommonVoice & VoxPopuli & MALACH \\
    \midrule
    W2V-ASRSpec (no LM)      & 7.29 & 11.28 & 18.93 \\
    \, + LM-ASRSpec & 6.12 & 10.51 & 18.45 \\
    \, + LM-ASRAll  & 5.34 &  9.78 & 16.67 \\
    \, + LM-C5      & 5.45 &  9.62 & 15.31 \\
    \bottomrule
  \end{tabular}
\end{table}
}

In the following sections, we experimented with all presented LMs but observed very similar trends, so we decided to report only results with LM-C5 since it either improved the model by correctly recognizing difficult domain-specific words or -- in the case of CommonVoice -- insignificantly deteriorated the results when compared to LM-ASRAll.

%\subsection{Window size and overlap}
%experiment s různou délkou okna a překryvu -> TSD
% Stačí udělat jenom na MALACHu, ostatní jsou krátké
%+ tabulka/graf

\subsection{Zero-shot transfer learning}
In the next experiment, we investigated the Wav2Vec's zero-shot transfer-learning ability between datasets, i.e. how well the model can transfer knowledge from one or more domains to an unobserved target domain with zero additional training. 

First of all, we cross-evaluated models fine-tuned on a single domain. These models were trained only from a single dataset.
Then, we tried the opposite setting, i.e. to fine-tune with all training datasets except for the one the model was evaluated on. Results with these out-of-domain models  (denoted as W2V-ood) correspond to a scenario where the model is fine-tuned from a large-scale ASR dataset (6 thousand hours) and subsequently used to recognize speech from an unobserved domain with zero additional training.
The zero-shot transfer results are shown in Tab.~\ref{tab:zeroshot}.

\setlength{\tabcolsep}{0.3em}
{\renewcommand{\arraystretch}{1.0}%
\begin{table}[htb]
  \caption{WER $[\%]$ for zero-shot transfer learning  evaluated on three Czech datasets. We are reporting results of models fine-tuned from single-domain training data (the first 3 models, each with and without LM, and all training data except for the in-domain data (W2V-ood).}
  \label{tab:zeroshot}
  \centering
  \begin{tabular}{lrrr}
    \toprule
     & CommonVoice & VoxPopuli & MALACH \\
     \midrule
    W2V-CommonVoice    & - & 15.62 & 26.57 \\		 		
    \, + LM-C5         & - & 13.81 & 22.33 \\
    W2V-VoxPopuli      & 29.94 & - & 33.37 \\		 	  	 		
    \, + LM-C5         & 22.57 & - & 28.04 \\	
    W2V-MALACH         & 19.21 & 16.89 & - \\	  	 			 		
    \, + LM-C5         & 11.63 & 13.36 & - \\ 
    \midrule
    W2V-ood            & 11.57 & 13.12 & 16.86 \\		
    \, + LM-C5         &  6.40 & 11.22 & 13.55 \\ 
    \bottomrule
  \end{tabular}
\end{table}
}

When comparing W2V-ood model with in-domain models (Tab.~\ref{tab:CTCLM}) for CommonVoice and VoxPopuli datasets, in-domain models are significantly better (with relative improvement about 15\%), which is the expected result. However, we observed a strange WER improvement when evaluating the W2V-ood model on the MALACH dataset. After some investigation, we found out, that it was caused by a mix of formal and colloquial Czech in transcripts. 
Formal Czech is the grammatically correct written form, while colloquial Czech is a spoken (common) form containing for example different suffixes (e.g. "mladý" vs. "mladej") or shortened word forms (e.g. "jsem" vs. "sem"). More details can be found in \cite{1306515}.
While in the training transcripts, the majority of the transcribed text is in colloquial Czech (i.e. transcribed exactly as spoken), the opposite is true for the test transcriptions. So, when the in-domain model was fine-tuned from the training MALACH data, it was forced to decode rather a colloquial Czech, which, however, conflicted with the test transcripts, causing a lot of word-substitution errors. For this reason, the out-of-domain model can perform significantly better on the MALACH dataset than the in-domain model.

\subsection{General vs. in-domain models}
\label{sec:general}
In the next experiment, we compared in-domain models (W2V-ASRSpec) with a general model fine-tuned from all available training data (i.e. both in-domain and out-of-domain) at once. Moreover, we were experimenting with a 2-phase fine-tuning procedure, in which the pretrained model is first fine-tuned from multi-domain data (same as the general model) and then fine-tuned again using only the in-domain data. In this approach, the model can benefit from large-scale out-of-domain data and, at the same time, the in-domain information is accentuated. 

\setlength{\tabcolsep}{0.4em}
{\renewcommand{\arraystretch}{1.0}%
\begin{table}[htb]
  \caption{WER $[\%]$ for differently fine-tuned models evaluated on three Czech datasets. We report results from models fine-tuned on all ASR data, only in-domain data, and a combination of both within a 2-phase fine-tuning procedure. We report all results without an LM and when decoded with LM-C5.}
  \label{tab:genASR}
  \centering
  \begin{tabular}{lrrr}
    \toprule
     & \hspace{-3em} CommonVoice & VoxPopuli & MALACH \\
    \midrule
    W2V-general          & 11.17 & 12.37 & 19.71 \\		
    \, + LM-C5         &  6.10 & 10.13 & 13.64 \\ 
    \midrule
    W2V-ASRSpec      & 7.29 & 11.28 & 18.93 \\
    \, + LM-C5      & 5.45 &  9.62 & 15.31 \\
    \midrule 
    W2V-general + ASRSpec       &  6.52 & 10.07 & 18.63 \\ 	
    \, + LM-C5         &  4.74 &  8.80 & 15.19 \\ 		
    \bottomrule
  \end{tabular}
\end{table}
}

Results are shown in Tab.~\ref{tab:genASR}.
Our suggested 2-phase fine-tuning performed significantly better than both the general and in-domain models. The only exception is the MALACH dataset, where in-domain models with LM again learned to decode colloquial Czech contrasting with test transcripts.

\subsection{Scaling up a batch size and updates}
So far, all models were fine-tuned with the default setting \cite{baevski2020wav2vec}, i.e.~80 thousand training steps with a batch size of about 27 minutes of audio corresponding to more than 5 epochs over all speech recognition data. Since we had large-scale ASR data, the natural question we asked was: Would it be beneficial to train the general model for a longer time? And with a larger batch?

To answer these questions, we fine-tuned the general model for a higher number of updates and/or with larger batch sizes, effectively increasing the number of training epochs over data. We fine-tuned with learning rate $8 \times 10^{-5}$ which we found to be better for large-scale datasets. Our results in Tab.~\ref{tab:ASRscale} show that
Wav2Vec speech recognizers scale very well with increasing the number of fine-tuning epochs over large datasets. For example, when we trained the model for 8x more epochs using a batch size of 108 minutes (4x larger than the default model) and 160 thousand update steps (2x more than default), the total error rate of end-to-end ASR decreased from 15.47\% to 13.33\%. Moreover, additional in-domain fine-tuning within the 2-phase procedure as described in Sec. \ref{sec:general} (model denoted as 40 epochs + ASRSpec) further improved the recognition. In the second phase of fine-tuning, we used the default settings since the underlying in-domain data were again small. 

\setlength{\tabcolsep}{0.6em}
{\renewcommand{\arraystretch}{1.2}%
\begin{table*}[t]
  \caption{WER $[\%]$ for scaling the fine-tuning epochs up. We show how the model performs when increasing the batch size (BS) and/or the number of updates (UP) by multiplying the default values. For each model, we show WER without LM and with LM-C5 for individual datasets, and total WER computed by aggregating numbers of words and errors over all 3 datasets.}
  \label{tab:ASRscale}
  \centering
  \begin{tabular}{lrrcrrcrrcrr}
    \toprule
     & \multicolumn{2}{c}{CommonVoice} & & \multicolumn{2}{c}{VoxPopuli} & & \multicolumn{2}{c}{MALACH} & & \multicolumn{2}{c}{TOTAL} \\
    \cline{2-3} \cline{5-6} \cline{8-9}  \cline{11-12} 
     & no LM & LM-C5 & & no LM & LM-C5 & & no LM & LM-C5 & & no LM & LM-C5 \\
    \midrule
    5 epochs (default)     & 11.17 & 6.10 & & 12.37 & 10.13 & & 19.71 & 13.64  & & 15.47 & 10.43 \\
    10 epochs (2xUP)       & 9.30  & 5.04 & & 11.00 &  9.42 & & 18.62 & 13.33 & & 14.07 & 9.79 \\
    10 epochs (2xBS)       & 9.23  & 4.82 & & 11.18 &  9.52 & & 19.05 & 13.60 & & 14.28 & 9.86 \\
    20 epochs (2xBS, 2xUP) & 8.49  & 4.59 & & 10.75 &  9.30 & & 17.99 & \textbf{12.98} & & 13.45 & \textbf{9.44}  \\
    20 epochs (4xBS)       & 8.39  & 4.62 & & 10.82 &  9.31 & & 18.86 & 13.89 & & 13.84 & 9.89 \\
    40 epochs (4xBS, 2xUP) & 7.68  & 4.29 & & 10.23 &  8.81 & & 18.52 & 13.73 & & 13.33 & 9.61 \\
    \, + ASRSpec        & \textbf{5.41}  & \textbf{3.80} & & \textbf{10.07} &  \textbf{8.80} & & \textbf{17.65} & 14.51 & & \textbf{12.10} & 9.82 \\
    \bottomrule
  \end{tabular}
\end{table*}
}

Also, our results show that models trained for the same number of epochs performed about the same no matter we multiplied the batch size or the number of updates. Results with the MALACH dataset with LM were again affected by the inconsistency of train and test texts leading sometimes to models preferring colloquial Czech over formal Czech and thus causing a large number of recognition errors.

\subsection{Comparison with other models}
Wav2Vec models are a very promising research area and we wanted to compare them also with existing ASR systems. 
As for the state-of-the-art results, we did not find any paper that reported WER on Czech CommonVoice data. 
%The only result we found is a self-reported $\text{WER}=7.27\%$\footnote{\url{https://huggingface.co/comodoro/wav2vec2-xls-r-300m-cs-250}} using version 8.0. 
For VoxPopuli, the best reported result is $\text{WER}=11.8\%$ from \cite{wang-etal-2021-voxpopuli} and for MALACH $\text{WER}=14.65\%$ from \cite{MALACH2}.

Since we did not develop an LVCSR system specifically for public CommonVoice and VoxPopuli datasets, we arranged zero-shot transfer learning conditions, under which the comparison can be made. We used the LVCSR system developed for real-time applications employed in live TV subtitling through respeaking \cite{LVCSR}. We compared this model with the W2V-ood model. Thus, both systems were trained for other domains and did not see any training examples (except for unlabeled data during pretraining of Wav2Vec). We slightly modified acronyms and multiwords in the reference to correspond with the texts used in the LVCSR (so the results are not directly comparable with other results in this paper). For both systems, we used the same LM (3-gram from all ASR transcripts). 

\setlength{\tabcolsep}{0.2em}
{\renewcommand{\arraystretch}{1.0}%
\begin{table}[htb]
  \caption{Comparison of LVCSR and Wav2vec models on three Czech datasets in terms of WER $[\%]$.}
  \label{tab:LVCSR1}
  \centering
  \begin{tabular}{lrrr}
    \toprule
     & CommonVoice & VoxPopuli & MALACH \\
    \midrule
    LVCSR          & 8.62 & 11.83 & 14.35 \\		
    Wav2Vec        & 5.75 &  9.01 & 12.93 \\
    \bottomrule
  \end{tabular}
\end{table}
}

For the MALACH dataset, we were comparing Wav2Vec models with the state-of-the-art LVCSR models developed specifically for this dataset. The LVCSR system was a CNN-TDNN LF-MMI with iVectors and sMBR criterion and a carefully curated 3-gram LM \cite{MALACH2}. We compared this model with our best-scoring Wav2Vec model, which is a general model fine-tuned from all ASR data (including MALACH) for 20 epochs. We used the same LM when decoding the output. Our results are summarized in Tab.~\ref{tab:LVCSR1}.

\section{Discussion}

Results in Tab.~\ref{tab:LVCSR1} clearly show that Wav2Vec with an LM-based beam search decoder is a robust ASR system, which can take advantage of large labeled and unlabeled datasets and successfully compete with state-of-the-art LVCSR systems. 

However, LVCSR systems have still some advantages over transformer-based systems. To mention a few of them: LVCSR system can be used for online real-time recognition with low latency; LVCSR system can easily handle irregular and difficult pronunciation of special words in a straightforward explicit way in the lexicon. Transformer-based ASR systems, on the other hand, are lexicon-free systems that can learn how to transcribe words without the need for explicit phonetic transcriptions, allowing any text (even huge corpora with vocabulary sizes in the order of tens of million words) to be used for language modeling in the Wav2Vec's decoder without any additional manual or engineering effort.

Results from our paper showed that the performance of the Wav2Vec-based ASR system can be significantly improved by: 
\begin{itemize}
    \item longer fine-tuning over the large-scale dataset, 
    \item adding LM into the beam search CTC decoder; we found the Common Crawl project to be a useful corpus for large-scale language modeling, 
    \item 2-phase fine-tuning procedure - multi-domain fine-tuning in the first phase followed by a second phase of fine-tuning using only data from the target domain.
\end{itemize}
Our best results with LM-C5 in Tab.~\ref{tab:ASRscale} are -- to our best knowledge -- new state-of-the-art results in all three datasets.

\section{Conclusions}
In this paper, we pretrained Czech monolingual audio transformers from a large dataset containing more than 80 thousand hours of unlabeled speech and released the model to the public. 
We presented a large palette of experiments with various fine-tuning setups with large-scale datasets (up to 6 thousand hours) evaluated on two public datasets (CommonVoice and VoxPopuli) and one extremely challenging dataset from the MALACH project.
Our results showed that monolingual Wav2Vec2 models are robust ASR systems, which can take advantage of large labeled and unlabeled datasets and successfully compete with state-of-the-art LVCSR systems. Moreover, our general transformer-based ASR systems proved to be good zero-shot learners when no training data are available for the target domain.
Our models scored new state-of-the-art results on all three tested datasets by a large margin.

\section{Acknowledgements}

% potřebujeme do nějakého projektu?

Computational resources were supplied by the project "e-Infrastruktura CZ" (e-INFRA CZ LM2018140 ) supported by the Ministry of Education, Youth and Sports of the Czech Republic.

\bibliographystyle{IEEEtran}

\bibliography{main}

\end{document}